\documentclass[final,5p, timesm twocolumn]{elsarticle}

\usepackage{amssymb}

\usepackage{times}
\usepackage{epsfig}
\usepackage{graphicx}
\usepackage{float}
\usepackage{amsmath}
\usepackage{amssymb}

\usepackage{tabu}
\usepackage{multirow}
\usepackage{makecell}

\usepackage{setspace}

\usepackage[switch]{lineno}
\usepackage[pagebackref=true,breaklinks=true,colorlinks,bookmarks=false]{hyperref}

\journal{Neurocomputing}

\begin{document}

\begin{frontmatter}

\title{Underwater Camouflage Object Detection Dataset}

\author{Feng Dong\textsuperscript{\rm 1,\rm 2}, Jinchao Zhu*\textsuperscript{\rm 3,1}\\
\textsuperscript{\rm 1} College of Artificial Intelligence, Nankai University, Tianjin, 300000, China\\
\textsuperscript{\rm 2} School of Finance, Tianjin University of Finance and Economics, Tianjin, 300000, China\\
\textsuperscript{\rm 3} Department of Automation, BNRist, Tsinghua University, 100089, China\\
{\tt\small jczhu@mail.nankai.edu.cn, dongfengdeyx@163.com, }}

\begin{abstract}
We have made a dataset of camouflage object detection mainly for complex seabed scenes, and named it UnderWater RGB\&Sonar, or UW-RS for short. The UW-RS dataset contains a total of 1972 image data. The dataset mainly consists of two parts, namely underwater optical data part (UW-R dataset) and underwater sonar data part (UW-S dataset).
\end{abstract}



\end{frontmatter}

\section{UW-RS dataset}
There are many datasets for salient object detection and camouflage object detection. The datasets for salient object detection include DUTS, DUT-OMRON, ECSSD, PASCAL-S, HKU-IS, etc. The datasets for camouflage object detection include CAMO, CHAMELEON, CPD1K, COD10K, NC4K, etc.
The image scenes of these datasets are very rich and challenging, and the scenes of camouflaging object detection datasets are particularly complex and diverse, such as field forest scenes, snowfield scenes, etc.

In order to promote the research of camouflage object detection in underwater environment, we have made an underwater dataset mainly for complex underwater scenes and named it UnderWater-RGB\&Sonar, abbreviated as UW-RS.The UW-RS dataset contains 1972 images in total.The dataset consists of two parts, namely, the underwater optical data section containing 1472 images (UW-R dataset) and the underwater sonar data section containing 502 images (UW-S dataset), where R and S represent the RGB image of visible light and the side-sweep sonar image respectively.
The UW-R dataset collates a large number of images containing underwater camouflage objects. The images and labels of the dataset are selected from the datasets CAMO, CHAMELEON, CPD1K, COD10K and NC4K.
The UW-S dataset contains a large number of side-sweep sonar images. The side-sweep sonar data are collected from the Internet, the open side-sweep sonar image classification dataset, \cite{2020-ACCESS-SonarClass}, the display images in literature, \cite{2017-Tcyber-RFM-SonarSeg, 2020-DNNGabor-Dec, 2009-OCEANS-unspvDet} and the images we collected through the sonar equipment.
The label of the side-scan sonar image is manually labeled by 5 professionals with knowledge of the side-scan sonar image.
Sound equipment is mostly used in the military. Because of the confidential problems, there is less open sonar image data.
UW-RS dataset is a very valuable and public dataset for side-scan sonar image segmentation.

\begin{figure*}[htb]
	\centering
	\includegraphics[width=2.05\columnwidth]{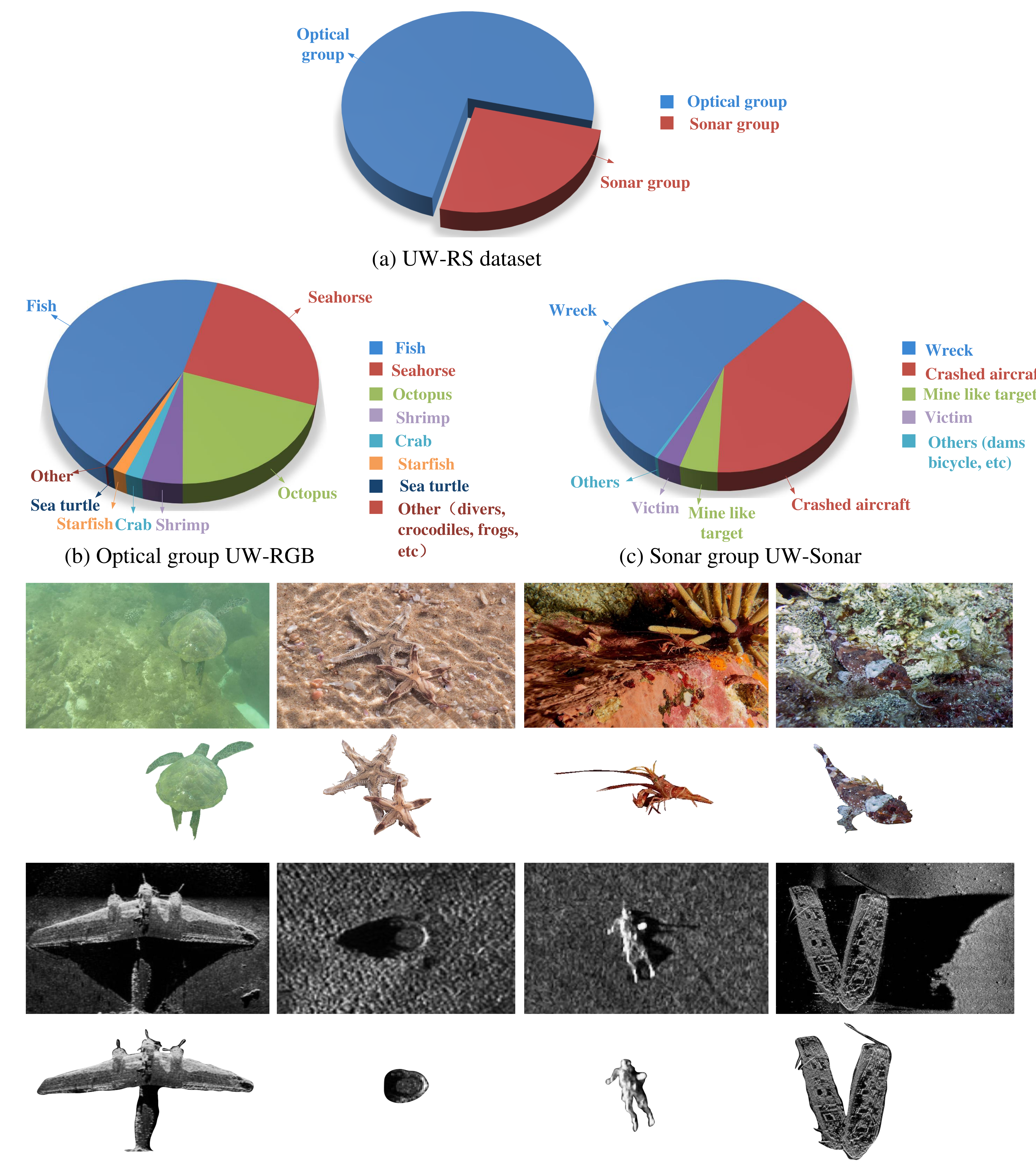} 
	\caption{Underwater dataset UW-RS for complex underwater scenes. (a) displays the proportion of optical group image to sonar group image. (b) displays the proportions of various objects in the optical group (UW-R) data. (c) displays the proportions of various objects in the Sonar Group (UW-S) data. Examples of images from some optical and sonar groups are shown at the bottom.
	}
	\label{UW1}
\end{figure*}

Fig. \ref{UW1} shows the composition of the UW-RS dataset, in which Fig. \ref{UW1}(a) shows the proportion of optical and sonar images in the dataset.
Fig. \ref{UW1}(b) shows the proportional relationships among various camouflaged object images contained in the optical group image.
Fish, sea dragon, octopus account for the largest proportion, followed by shrimp, crab, starfish and turtles.
A small number of images with disguised objects such as divers, crocodiles, frogs are combined into other categories.
Fig. \ref{UW1}(c) shows the proportional relation of various object images contained in the sonar group.
The largest proportion were wreck and crashed aircrafts, followed by mine like targets, victims and others.
A small number of images with objects such as dams, bicycles are combined into other categories.
Below Fig. \ref{UW1} are some image examples of the optical group and some image examples of the sonar group in the UW-RS dataset.
To show the foreground more clearly, we use tags to extract the foreground and adjust the dimensions for better presentation.
It can be seen from the illustration that the boundary of the object is very fine, which means that the label marking is meticulous and accurate.

\begin{figure*}[htb]
	\centering
	\includegraphics[width=2.05\columnwidth]{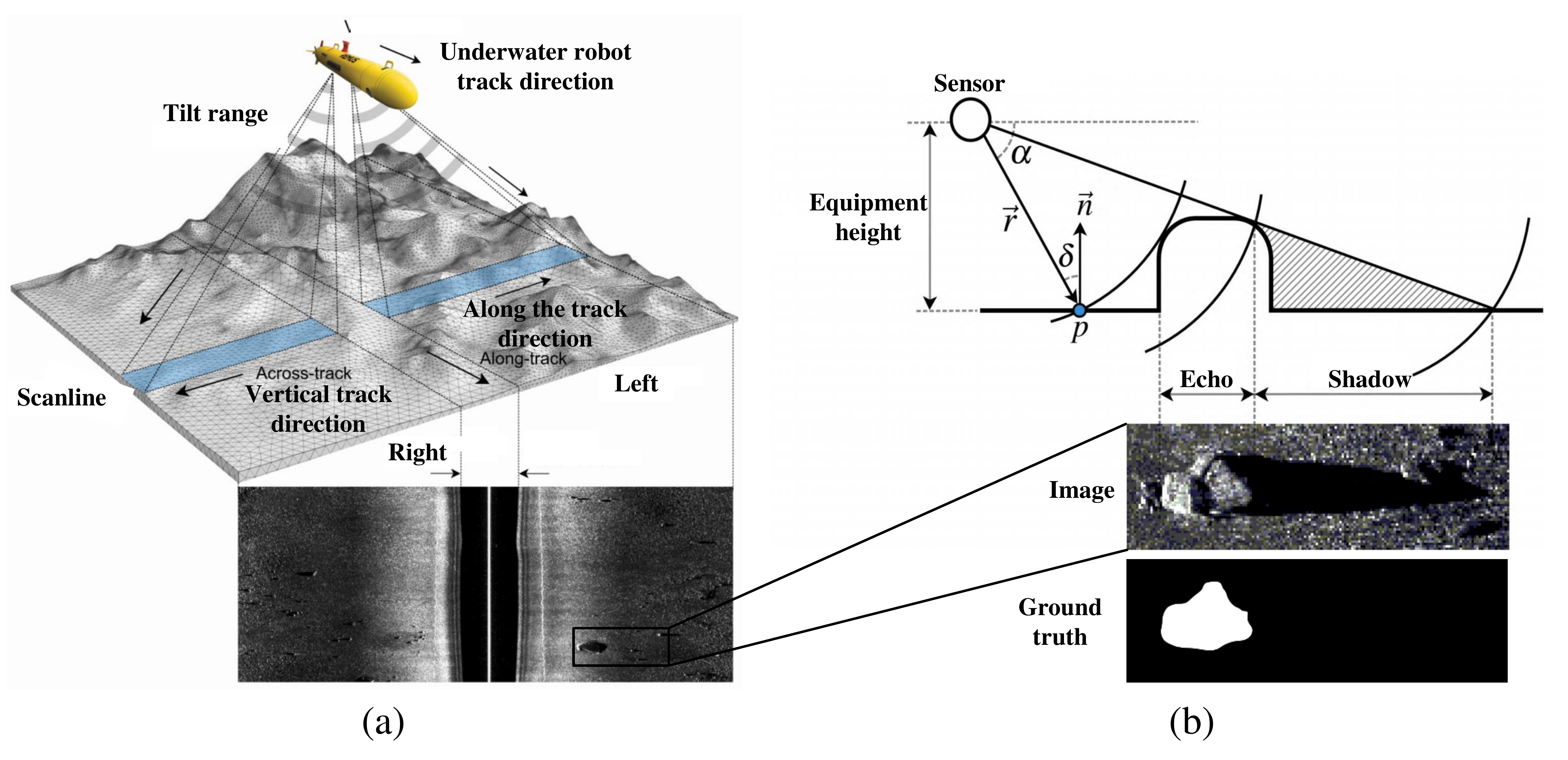}
	\caption{Introduce the principle of side-sweep sonar and display of side-sweep sonar image and label.
	}
	\label{UW2}
\end{figure*}

Fig. \ref{UW2}  shows the acquisition process of side-scan sonar image. Fig. \ref{UW2}(a) shows the underwater operation of an underwater vehicle, which carries a side-sweeping sonar device that collects information about underwater scenes.
Fig. \ref{UW2}(b) shows in detail the information of the echo and shadow areas perceived by the sensor.
Below Fig. \ref{UW2}(b) shows the side-sweep sonar image and its corresponding tags from the UW-RS dataset.

\bibliographystyle{unsrt}
\bibliography{nkthesis}  
\end{document}